\pdfoutput=1

\documentclass[11pt]{article}

\usepackage[final]{acl}

\usepackage{times}
\usepackage{latexsym}
\usepackage{amsfonts}
\usepackage{todonotes}
\usepackage{algpseudocodex}

\usepackage[T1]{fontenc}

\usepackage[utf8]{inputenc}

\usepackage{microtype}

\usepackage{inconsolata}
\usepackage{booktabs}
\usepackage{amsmath}
\usepackage[nameinlink,capitalize,noabbrev]{cleveref}

\usepackage{xparse}

\usepackage{tikz}
\usetikzlibrary{arrows.meta}
\usetikzlibrary{calc}
\usetikzlibrary{colorbrewer}
\usetikzlibrary{fit}
\usetikzlibrary{matrix}
\usetikzlibrary{positioning}

\usepackage{pgfplots}
\usepackage{pgfplotstable}
\pgfplotsset{compat=1.17}
\pgfplotsset{
        modern/.style={
                enlargelimits=false,
                separate axis lines,
                semithick,
                axis x line*=bottom,
                axis x line shift=10pt,
                axis y line*=left,
                axis y line shift=10pt,
                every axis/.append style={thick},
                tick style={thick, black},
                tick align=outside,
        },
        points/.style={only marks, mark options={draw=pblue, fill=pblue, scale=0.5}},
        regression/.style={no markers, porange, very thick}
}
\pgfkeys{/pgf/number format/.cd,1000 sep={\,}}

\usepackage{xcolor}

\definecolor{pblue}{HTML}{377eb8}
\definecolor{porange}{HTML}{ff7f00}
\definecolor{pgreen}{HTML}{4daf4a}
\definecolor{ppink}{HTML}{984ea3}

\NewDocumentCommand\head{}{\ensuremath s}
\NewDocumentCommand\tail{}{\ensuremath e}

\title{Compositional Generalization with Grounded Language Models}

\author{Sondre Wold, Étienne Simon, Lucas Georges Gabriel Charpentier, \\ {\bf Egor V. Kostylev}, {\bf Erik Velldal}, {\bf Lilja Øvrelid} \\ University of Oslo}

\begin{document}
\maketitle
\begin{abstract}
Grounded language models use external sources of information, such as knowledge graphs, to meet some of the general challenges associated with pre-training. By extending previous work on compositional generalization in semantic parsing, we allow for a controlled evaluation of the degree to which these models learn and generalize from patterns in knowledge graphs. We develop a procedure for generating natural language questions paired with knowledge graphs that targets different aspects of compositionality and further avoids grounding the language models in information already encoded implicitly in their weights. We evaluate existing methods for combining language models with knowledge graphs and find them to struggle with generalization to sequences of unseen lengths and to novel combinations of seen base components. While our experimental results provide some insight into the expressive power of these models, we hope our work and released datasets motivate future research on how to better combine language models with structured knowledge representations.
\end{abstract}

\section{Introduction}
Language models (LMs) acquire latent representations of text through their pre-training. One drawback of this paradigm is the implicit encoding of knowledge in a large parameter space, which can lead to factual hallucinations in model outputs. To reduce this effect, there has recently been a widespread interest in what is sometimes referred to as \textit{grounded language models}, which are LMs grounded in external information sources, such as knowledge graphs (KGs). These approaches typically attempt to enable reasoning over KGs by combining the LM outputs with that of a separate graph encoder, such as a graph neural network~\citep[GNN;][]{lin-etal-2019-kagnet, greaselm, sun-etal-2022-jointlk, yasunaga2022deep, Yu_Zhu_Yang_Zeng_2022}. It can be difficult to determine to what extent grounded models exhibit any structured reasoning, or what precisely would constitute reasoning in this context, as the information contained in these KGs often targets the same domain as the corpora used for LM pre-training. One example of this is the overlap in information between Wikidata, often used as the KG in previous work on grounded LMs, and Wikipedia, which is a common corpus for pre-training.

Central to human reasoning is the ability to form novel combinations from seen components, often referred to as compositional generalization. Studies on compositionality have a long tradition within many fields of study, such as linguistics \citep{partee1995lexical}, philosophy \citep{pagin2010compositionality}, and cognitive science \citep{churchland1990cognitive}, and determining to what degree neural networks capture compositionality is a long-standing research problem in machine learning, with debates going back several decades \citep{fodor1988connectionism, smolensky1987connectionist}. Recent studies in NLP have primarily used semantic parsing of synthetic utterances as a testbed for investigating to what extent neural networks combine syntactic elements to form new complex expressions \citep{hosseini-etal-2022-compositional}. One of the most well-known examples of this is from \citet{lake2018generalization}, who found \textit{seq2seq} models to struggle both with systematic compositional skills and generalization to longer sequences than seen during training -- abilities that are often taken to be central to compositional generalization. Some of these shortcomings were recently mitigated using a meta-learning framework \citep{lake2023comp}, but it is still not clear under what circumstances neural networks can consistently demonstrate compositional generalization, or if they can at all \citep{dziri2024faith}.

In this work, we investigate to what extent grounded language models exhibit compositional generalization by combining two orthogonal research directions: compositionality in neural networks and the grounding of LMs in KGs.

Using question answering as a task, we train models to recognize the relation between certain patterns in a synthetic KG and natural language questions. We develop a data generation procedure that targets three aspects of compositionality from \citet{hupkes2020compositionality}: \textit{substitutivity}, \textit{productivity}, and \textit{systematicity}. In our experiments, we evaluate LMs that interact with KGs through the use of a GNN, an approach previously proven effective for tasks such as question answering \citep{lin-etal-2019-kagnet, greaselm, sun-etal-2022-jointlk} and pre-training \citep{yasunaga2022deep}.

Our work targets two key limitations from the literature: Firstly, by extending existing work on compositional generalization to question answering over KGs, we can assess the capabilities of these models for structured reasoning over natural language questions of varying complexities; and secondly, by using a synthetic knowledge graph we do not run the risk of grounding a language model in a knowledge source containing information that already exists implicitly in the models' weights. The contribution of our work is three-fold: \emph{i)}~We provide the first experimental study on the expressive power of grounded LMs with respect to compositional generalization \emph{ii)}~We develop a procedure for generating dataset samples that target three theoretically motivated aspects of compositionality, and \emph{iii)}~We release a series of generated datasets using this procedure, together with all code and data related to our experiments, which can be used to benchmark grounded language models in a controlled environment.%
\footnote{\url{https://github.com/ltgoslo/text-graph-generalization}}

\section{Compositionality}\label{sec:overview_motivation}
In what follows we discuss three aspects of compositionality that have been established as tests of compositionality in previous work \citep{hupkes2020compositionality}: \textit{substitutivity}, \textit{systematicity}, and \textit{productivity}. We also describe how we ground these notions to be applicable for our setting, where data consists of both text and KGs. We refer an interested reader to the overview by \citet{sep-compositionality} for a more extensive introduction to these terms.

\begin{figure}[t]
        \centering
        \begin{tikzpicture}[
                entity/.style={fill=pblue, text=white, minimum height=8mm, rounded corners=2mm},
                link/.style={-{Latex}, rounded corners=1mm},
                relation/.style={above, rotate=90},
        ]
        \small
        \matrix[column sep=5mm, row sep=20mm] {
                \node[entity] (professor7) {Professor7}; & & \node[entity] (paper2) {Paper2}; \\
                \node[entity] (course31) {Course31};     & & \node[entity] (professor2) {Professor2}; \\
                \node[entity] (gradstudent9) {GradStudent9}; & \node[entity] (gradstudent3) {GradStudent3}; & \node[entity] (gradstudent5) {GradStudent5}; \\
        };
        \draw[link] (professor7) to node[relation] {\textsc{Teaches}} (course31);
        \draw[link] (professor2) to node[relation] {\textsc{advisor}} (gradstudent5);
        \draw[link] (professor2) to node[relation] {\textsc{pub.author}} (paper2);
        \draw[link] (course31) to node[relation] {\textsc{hasStudent}} (gradstudent9);
        \draw[link] (course31) -| node[pos=0.75, relation] {\textsc{hasStudent}} (gradstudent3.140);
        \draw[link] (professor2) -| node[pos=0.75, relation] {\textsc{advisor}} (gradstudent3.40);
        \coordinate (right) at ($(gradstudent5.east)+(4mm, 0)$);
        \draw[link] (gradstudent5) -- (right) -- node[relation] {\textsc{pub.author}} (paper2-|right) -- (paper2);
\end{tikzpicture}
        \caption{An example KG illustrating the relationship between entities from the LUBM inventory.}
        \label{fig:example_graph}
    \end{figure}
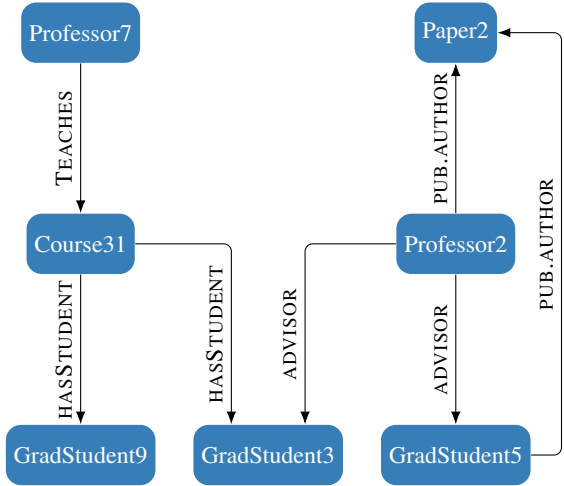

\paragraph{Substitutivity}
One of the most well-known definitions of compositionality is the \textit{principle of compositionality} from \citet{partee1995lexical}: ``The meaning of a compound expression is a function of the meanings of its parts and of the way they are syntactically combined''. Closely related to this principle is the notion of \textit{substitutivity}, which states that any change to a complex expression that maintains the meaning of individual parts also maintains the meaning of the whole expression. 
On the one hand, in the context of natural languages, this definition is not without controversy. Languages often have elements that do not adhere strictly to such a principle, for example through the use of idioms \citep{dankers-etal-2022-paradox}. On the other hand, in our setting, we can exploit the observation that if an atomic expression in natural language has a node in a graph as its reference, then we can substitute this reference with a compounded expression referencing the same node without changing the meaning of the whole expression. As the principle of compositionality is not committed to any specific theory of semantics, we can use \textit{reference} as the qualification for semantic equivalence, a theory sometimes referred to as the \textit{compositionality of reference} \citep{sep-compositionality}. For example, given the KG in \cref{fig:example_graph}, we can ask ourselves whether ``Paper2 was authored by Professor2''. We can then replace ``Professor2'' with a compounded reference, such as ``the supervisor of GraduateStudent5'', which gives rise to the question: ``Was Paper2 authored by the supervisor of GraduateStudent5?'' Following the compositionality of reference, we do not change the meaning of the overall question, since we still enquire about the same node in the graph, even though we have substituted a 1-hop reference with a 2-hop reference. Consequently, by constructing highly compositional questions in natural language, we can test whether or not neural networks can learn the mapping between constituents in text and paths in a graph. 

\paragraph{Productivity}
Closely related to substitutivity, \textit{productivity} refers to the ability to construct seemingly infinite constructions using a finite capacity. As \citet{hupkes2020compositionality} point out, this view on the open-endedness of language is typically associated with generative linguistics, and in the context of neural networks, the most direct way of testing for this ability is to test models on generalization to sequences of lengths not seen during training. In our case, we can train models on questions that require reasoning over $k$-hop paths, for several $k$, in a graph, and then test on $j$-hop paths, for various $j$ different from the $k$'s used in training. 

\paragraph{Systematicity}
One of the most tested properties of compositionality is \textit{systematicity}, which was popularized as a term by  \citet{fodor1988connectionism}. Systematicity can loosely be described as the ability to combine seen parts and rules to form new combinations, an ability that has a long tradition within linguistics. In the context of neural networks, tests of systematicity typically involve exposing models to a finite set of items and combinations during training and then evaluating the same models on a test set containing the same set of items, but combined in ways not seen during training. \citet{lake2018generalization} use the example of a person that knows the meaning and usage of words such as ``twice,'' ``and,'' and ``again,'' which upon learning the new verb ``to dax'' can generalize to new constructions such as ``dax twice and then dax again,'' something they find neural networks to struggle with. Recently, there have also been some attempts to formalize and quantify systematic generalization in neural networks \citep{keysers2020measuring, ram2023how}. 
In our setting, we can use the typed relations of the edges in a graph, such as the one illustrated in \cref{fig:example_graph}, to test for systematicity. If a model has seen the relations \textsc{teaches}, \textsc{hasStudent}, and \textsc{advisor} in numerous 3-hop paths, but never in this order, a display of systematicity would require generalization to this unseen order.

\section{Dataset Generation}\label{sec:dataset}
\subsection{Overview}
To target the compositional abilities specified above, we generate several datasets consisting of pairs of  KGs and yes/no questions in natural language, all of which are about the information in these graphs. This section explains the properties of these datasets and how they are created.

A KG over a set $\mathcal{T}$ of \emph{types} and a set $\mathcal R$ of \emph{relations} is a directed labeled multi-graph ${(\mathcal{V}, \mathcal{L}, \mathcal{E})}$ with nodes $\mathcal{V}$, called \emph{entities} in this context, set $\mathcal{L}$ of node labelings, each of which is of the form $(v, t)$ where $v$ is an entity in $\mathcal{V}$ and $t$ a type in $\mathcal{T}$, and set $\mathcal{E}$ of edges,
each of which is of the form $(v, r, v')$, where $v$ and $v'$ are entities from $\mathcal V$, and $r$ is a relation from $\mathcal R$.
In this work, we only concentrate on graphs from the Lehigh University Benchmark \citep[LUBM;][]{guo2005lubm}, which is a synthetic KG benchmark popular for evaluating various aspects of knowledge graph repositories. In particular, we exploit LUBM's KG generator of customizable and repeatable synthetic data, which can produce KGs of a user-specified size describing a fictitious university domain and committing to a realistic ontology over sets $\mathcal T_{\textsf{LUBM}}$ and $\mathcal R_{\textsf{LUBM}}$ of university-related types and relations. We generate several graphs with an average number of 685 entities, each labeled by a single type, and 4\,949 edges, each labeled by a single relation.%
\footnote{The ontology is specified in Web Ontology Language OWL (\url{https://www.w3.org/OWL/}).}

Each of the natural language questions targets two entities in the corresponding generated LUBM graph $\mathcal{G}$ and a chain of relations from $\mathcal R_{\textsf{LUBM}}$ and inverses of such relations that is to hold between these entities. 
In particular, for a number of hops (i.e., chain length) $k$, the questions are semi-manually created by first mapping each unique $k$-hop combination (i.e., chain of length $k$) of relations from a selected subset of $\mathcal{R}_{\textsf{LUBM}}$ and their inverses to a natural language template with two parameters, and then substituting these 
parameters by two entities from $\mathcal{V}$ as arguments. As a result, the answer to the question is \textsc{true} if there exists a path between the two entities in $\mathcal{G}$ as prompted by the $k$-hop combination of the question template, and \textsc{false} otherwise. We create templates for 2, 3, and 4 hops, with 5 relations from $\mathcal{R}_{\textsf{LUBM}}$ and their inverses as $\mathcal{R}_{\textsf{LUBM}}'$. This makes for over 280 unique templates (while the number of questions depends on $\mathcal{G}$). 
To illustrate our dataset creation we concentrate on an example and a discussion of a key step, sampling of positive and negative question--graph pairs, in the next two subsections. We refer to \cref{sec:appendix_dataset_generation} for further details.

\subsection{Example}
Given the KG in \cref{fig:example_graph}, we can create the following templates, for an increasing number of hops, where the ``$-$'' superscript denotes the inverse of the relation:
\begin{description}
    \item[2 hops:] \textsc{[teaches, takesCourse$^-$]} $\rightarrow$ Does \head{} teach a course that has a student named \tail{}?
    \item[3 hops:] \textsc{[teaches, takesCourse$^-$, advisor]} $\rightarrow$ Does \head{} teach a course that has a student supervised by \tail{}?
    \item[4 hops:] \textsc{[teaches, takesCourse$^-$, advisor, publicationAuthor]} $\rightarrow$ Does \head{} teach a course that has a student supervised by an author of \tail{}?
\end{description}
To create questions from these templates that we label \textsc{true} we can use \textsc{Professor7} and \textsc{GradStudent9} as arguments substituting parameters \head{} and \tail{}, respectively, in the 2-hop case, \textsc{Professor7} and \textsc{Professor2} for the 3-hop case, and \textsc{Professor7} and \textsc{Paper2} for the 4-hop case. Creating examples labeled as \textsc{false} for these templates then involves sampling two entities to use as \head{} and \tail{} that are \emph{not} connected by the chains of relations, but are similarly connected in the graph as the entities in the positive examples. We describe this negative sampling in the next section.

\subsection{Sampling Procedure}
As mentioned, we create each question--graph pair by sampling pairs of entities and the relations on the edges that are to hold between them in the graph $\mathcal{G}$. The key requirement to this procedure is to guarantee that the $k$ relations and their inverses in every sampled $k$-hop combination are as likely to be labeled \textsc{true} as they are to be labeled \textsc{false}; failing to do so introduces distributional bias. This means that our sampling procedure must ensure two important properties: \emph{i)} it must be impossible for a text-only model to associate a question template (i.e., all questions produced from this template) to the \textsc{true} or \textsc{false} label, and \emph{ii)} it must be impossible for a graph-only model to associate the presence of a specific set of relations in graph $\mathcal{G}$ with a specific label. 

We ensure property \emph{i)} by creating the same number of questions labeled \textsc{true} as \textsc{false} for every unique $k$-hop combination of relations. We ensure \emph{ii)} by sampling $k$-hop relation paths from $\mathcal{G}$ symmetrically: We first sample i.i.d.~two pairs of different entities, $\head_+$, $\tail_+$ and $\head_-$, $\tail_-$, such that $\head_+$ and $\head_-$ are of the same entity type, and $\tail_+$ and $\tail_-$ are also of the same entity type. We then identify a $k$-hop combination over $\mathcal R_{\textsf{LUBM}}'$ that exists between $\head_+$ and $\tail_+$ but not between $\head_-$ and $\tail_-$, as well as verify that there exists another (different) $k$-hop combination over $\mathcal R_{\textsf{LUBM}}'$ between $\head_-$ and $\tail_-$. If all this succeeds, the pair $\head_+$, $\tail_+$ with the question corresponding to the template becomes a positive example, and the pair $\head_-$, $\tail_-$ with the same template a negative one. Further details of the sampling procedure can be found in \cref{sec:appendix_dataset_generation}.

\section{Modeling}

For all experiments, each data point includes a natural language question $q$ and a graph $\mathcal{G}$. Following previous work on combining language models with KGs \citep{lin-etal-2019-kagnet, zhang-etal-2020-grounded, yasunaga-etal-2021-qa, greaselm, kaur-etal-2022-lm, sun-etal-2022-jointlk, yasunaga2022deep}, we use separate encoders for $q$ and $\mathcal{G}$ to obtain representations of both modalities. \cref{sec:encoders} outlines these encoders and \cref{sec:full_models} how we combine them to produce the final output prediction $\hat{y} \in \{\textsc{true}, \textsc{false}\}$.

\subsection{Encoders}\label{sec:encoders}

\paragraph{Text encoder}
We obtain representations of $q$ by using the special sentence representation token at the last layer of a pre-trained language model, as per \citet{devlin-etal-2019-bert}: $z_{\text{text}} = \operatorname{BERT}(q)$.

\paragraph{Graph encoder}
As a graph encoder, we use the relational graph convolution model, \textsc{R-GCN}, from \citet{schlichtkrull2018modeling}:
\begin{equation*} \label{eq_rgcn}
    h_i^{\ell+1} = \sigma \Bigg( \sum_{r \in \mathcal{R}} \sum_{j \in \mathcal{N}^r_i} \frac{1}{\mid \mathcal{N}^r_i \mid} W_r^{\ell}h_j^{\ell} + W_0^{\ell}h_i^{\ell} \Bigg),
\end{equation*}
where $h_i^{\ell}$ is the hidden state of a node $i$ in layer $\ell$, $\sigma$ is the ReLU function and $\mathcal{N}^r_i$ is the set of neighborhood vertices of node $i$ under relation $r$.

\subsection{Final models}\label{sec:full_models}
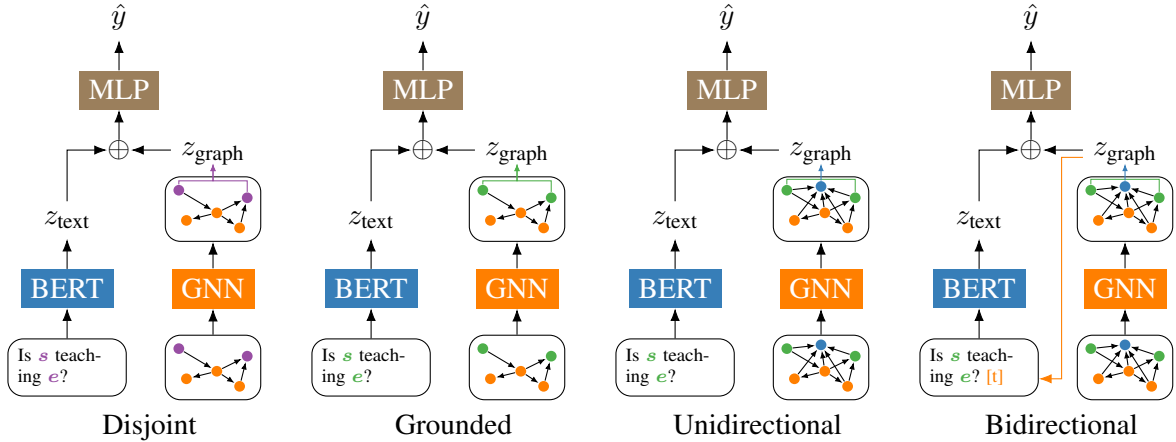
\begin{figure*}[h!]
    \centering
    \begin{tikzpicture}[
                vertex/.style={circle, fill=porange, inner sep=0.5mm},
                arc/.style={-{Latex[scale=0.6]}},
                flow/.style={-{Latex}},
                input/.style={draw, rounded corners=2mm}
        ]
        \NewDocumentCommand\tgraph{m m m}{
                \node[vertex, fill=#3] (#1 v1) at (#2) {};
                \node[vertex] (#1 v2) at ([yshift=-3mm, xshift=5mm]#1 v1) {};
                \node[vertex] (#1 v3) at ([yshift=-1mm, xshift=-4mm]#1 v2) {};
                \node[vertex] (#1 v4) at ([yshift=-2mm, xshift=3mm]#1 v2) {};
                \node[vertex, fill=#3] (#1 v5) at ([yshift=2mm, xshift=4mm]#1 v2) {};
                \draw[arc] (#1 v1) -- (#1 v2);
                \draw[arc] (#1 v2) -- (#1 v3);
                \draw[arc] (#1 v2) -- (#1 v4);
                \draw[arc] (#1 v2) -- (#1 v5);
                \draw[arc] (#1 v4) -- (#1 v5);
                \node[input, fit=(#1 v1)(#1 v2)(#1 v3)(#1 v4)(#1 v5), inner sep=1mm] (#1) {};
        }

        \foreach \n/\d in {Disjoint/0, Grounded/4, Unidirectional/8, Bidirectional/12}{
                \tgraph{input graph}{\d, 0}{\ifnum\d<2 ppink\else pgreen\fi}

                \node[input, text width=13mm, left=5mm of input graph, font=\scriptsize] (input text) {\raggedright Is \textcolor{\ifnum\d<2 ppink\else pgreen\fi}{\(\boldsymbol{\head}\)} teaching \textcolor{\ifnum\d<2 ppink\else pgreen\fi}{\(\boldsymbol{\tail}\)}?\ifnum\d=12\relax~\textcolor{porange}{[t]}\fi};

                \node[fill=pblue, text=white, above=4mm of input text] (bert) {BERT};
                \node[fill=porange, text=white] (gnn) at (bert-|input graph) {GNN};

                \draw[flow] (input graph) -- (gnn);
                \draw[flow] (input text) -- (bert);

                \node[above=4mm of bert] (z text) {\(z_\text{text}\)};
                \tgraph{output graph}{\d, 21mm}{\ifnum\d<2 ppink\else pgreen\fi}
                \node[above=1.5mm of output graph, inner ysep=0mm] (z graph) {\(z_\text{graph}\)};

                \draw[flow] (bert) -- (z text);
                \draw[flow] (gnn) -- (output graph);

                \coordinate (tmp) at ([yshift=-0.6mm]output graph.north);
                \coordinate (out) at ([yshift=0.2mm]z graph.south);

                \ifnum\d<6
                        \draw[arc, \ifnum\d<2 ppink\else pgreen\fi] (output graph v1) |- (tmp) -- (out);
                        \draw[arc, \ifnum\d<2 ppink\else pgreen\fi] (output graph v5) |- (tmp) -- (out);
                \fi


                \coordinate (tmp) at ([yshift=-0.4mm]output graph.north);

                \ifnum\d>6
                        \node[vertex, fill=pblue] (input graph u) at ([yshift=3.5mm, xshift=-0.5mm]input graph v2) {};
                        \node[vertex, fill=pblue] (output graph u) at ([yshift=3.5mm, xshift=-0.5mm]output graph v2) {};
                        \foreach \v in {1,...,5}{
                                \draw[arc] (input graph v\v) -- (input graph u);
                                \draw[arc] (output graph v\v) -- (output graph u);
                        }
                        \draw[pgreen] (output graph v1) |- (tmp);
                        \draw[pgreen] (output graph v5) |- (tmp);
                        \draw[arc, pblue] (output graph u) -- (out);
                \fi
                
                \coordinate (graph inter) at ([yshift=-1.5mm]$(input text.east)!0.6!(input graph.west)$);
                \ifnum\d>10
                        \draw[flow, porange] (z graph.185) -| (graph inter) -- (input text.east|-graph inter);
                \fi

                \coordinate (tmp) at (z text|-z graph.175);
                \node[inner sep=0mm] (concat) at ($(tmp)!0.5!(z graph.175)$) {\(\oplus\)};
                \draw[flow] (z graph.175) -- (concat);
                \draw[flow] (z text) |- (concat);

                \node[fill=pblue!50!porange, text=white, above=4mm of concat] (mlp) {MLP};
                \draw[flow] (concat) -- (mlp);

                \node[above=4mm of mlp] (y) {\(\hat{y}\)};
                \draw[flow] (mlp) -- (y);

                \node at ([yshift=-4mm]input text.south -| graph inter) {\strut \n};
        }
\end{tikzpicture}
    \caption{Overview of the different approaches to combining the two encoders. \textcolor{ppink}{Purple} nodes are initialized at random while \textcolor{pgreen}{green} and \textcolor{pblue}{blue} nodes are initialized using a frozen BERT model.}
    \label{fig:model_overview}
\end{figure*}
Inspired by previous work on combining text and graph encoders, we experiment with different approaches to integrating the two modalities and apply them using the above-mentioned encoders and our experimental setup. \cref{fig:model_overview} provides an overview of the differences between these approaches.

\paragraph{Disjoint}
The simplest way of having the two modalities interact is to initialize and run both encoders separately before conjoining their outputs in a final classification layer. This keeps the encoders disjoint, essentially making a ``two-tower architecture'' -- conceptually similar to approaches such as \citet{wang2019improving}.

To get a final representation of the graph, $z_{\text{graph}}$, we take the embedding of the head and tail node targeted by the question after $L$ rounds of message passing: $z_{\text{graph}} = h_{\head}^{L} \oplus h_{\tail}^{L}$,
where \head{} and \tail{} are the indices of the head and tail node and $\oplus$ is vector concatenation. We initialize these nodes by creating two distinct embeddings, $E_\head$ and $E_\tail$, with $E_\head, E_\tail \sim \mathcal{N}(0, 1)$. We keep these static throughout training and testing and use them for all head and tail nodes targeted by the questions. The initialization of remaining nodes in each graph is discussed in more detail in \cref{sec:node_inits}.

For the question, we take the output of the text encoder without any modifications. The final output prediction is then obtained by the following: $f(z_{\text{text}} \oplus z_{\text{graph}})$, where $f$ is a two-layered MLP with a ReLU activation function. 

\paragraph{Grounded}
For this model, we concatenate the original question string from $q$ with both the head and tail entity strings, respectively, using a separation token. This is then used as input to the text encoder, obtaining a representation of the head and tail nodes:
\begin{align*}
    h_\head^0 & = \operatorname{BERT}(q~\texttt{<SEP>}~\text{head surface form}),\\
    h_\tail^0 & = \operatorname{BERT}(q~\texttt{<SEP>}~\text{tail surface form}).
\end{align*}
These are then used to initialize the embeddings for the nodes pertaining to these entities in the graph. This is done in a pre-processing step before the text encoder is fine-tuned on our datasets. The final output prediction is calculated in the same way as with the disjoint model. 

\paragraph{Unidirectional}
In this model, we add a unidirectional interaction between the encoders, inspired by works such as \citet{yasunaga-etal-2021-qa}. For each question--graph pair, $(q, \mathcal{G})$, we introduce a new special context node, $u$, and connect $u$ to each node in $\mathcal{G}$. This special node is connected using two new relations, one for each direction, and its embedding is initialized using the text encoder representation of the question, $\operatorname{BERT}(q)$. In \cref{fig:model_overview}, $u$ is marked as a blue node. For each $\mathcal{G}$, we perform global pooling over the hidden states:
\begin{equation*}
    z_{\text{graph}} = \frac{1}{|\mathcal{V'}|} \sum_{v\in\mathcal{V'}} h^{L}_v, 
\end{equation*}
where $\mathcal{V'}$ is the set containing the head, tail, and special context nodes. We do not consider the set of all nodes due to the zero-initialization of the remaining nodes in $\mathcal{G}$ (see \cref{sec:node_inits}). The head and tail nodes are initialized as in the grounded model, using the text encoder representations. We also take the embedding of the special node at the last layer, $h^{L}_u$, as a representation of $\mathcal{G}$. The final output prediction is then given by: $f(z_{\text{text}} \oplus z_{\text{graph}} \oplus h^L_u)$, where $f$ is a two-layered MLP with a ReLU activation function.

\paragraph{Bidirectional}
We also develop a model inspired by the interaction approach used in works such as \citet{greaselm} and \citet{yasunaga2022deep} to allow for a bidirectional interaction between the two encoders. We keep the special context node $u$ as described in the unidirectional model, but we add a special interaction token, $t$, to the question $q$. $t$ is set to be the representation of the special context node $u$ after $L$ layers of message passing: $t = h^{L}_u$. This is done after the embedding of $q$ in the text encoder. We produce the final output prediction using the same method as in the unidirectional model.

\paragraph{Baseline}
We use a sole graph encoder as a baseline. Theoretically, it should not be possible to solve the tasks without having access to both the text and the graph. By using a graph-only system as a baseline we therefore get both a sanity check and an upper bound on the noise produced by the sampling. As the natural language questions are not sampled, but rather constructed deterministically from the paths sampled from our graphs, one positive and one negative, the performance of a text-only encoder will by design be equal to a random classifier. The output of the baseline is calculated the same way as for the disjoint model, taking the representations of the head and tail entity.

\subsection{Node initialization}\label{sec:node_inits}
Except for the head and tail nodes targeted by each question, and the special context node $u$, all nodes are initialized as zero-vectors.
This means that we leak the position of the head and tail entity in $\mathcal{G}$ for each sample in the dataset, making any pruning of $\mathcal{G}$ redundant. It is often necessary to reduce the computational complexity of combining language models and large knowledge graphs using mechanisms such as subgraph retrieval and entity linking, but these steps are also known to be error-prone \citep{wold-etal-2023-text}. This initialization ensures that we further isolate the application of compositional generalization rules, as opposed to evaluating the performance of a pipeline model with numerous error-prone components.

\section{Experiments and results}\label{sec:exp_res}
\begin{table}[ht]
\centering
\footnotesize
    \begin{tabular}{@{}c@{\hspace{2em}}cc@{}}
    \toprule
    \textbf{Hops} & \textbf{\#relation-paths} & \textbf{N} \\
    \midrule
    $2$ & $\hphantom{0}23$ & $1043^{\pm1941}$  \\
    $3$ & $\hphantom{0}73$ & $\hphantom{0}328^{\pm449\hphantom{0}}$  \\
    $4$ & $188$ & $\hphantom{0}127^{\pm260\hphantom{0}}$  \\
    \bottomrule
    \end{tabular}
    \caption{The mean number of occurrences of the relation paths across 24\,000 generated graphs per hop $k$.}
    \label{tab:data_statistics}
\end{table}

In this section, we test our suite of models on data specifically targeting substitutivity (\cref{sub_sec:subst}), productivity (\cref{sub_sec:prod}), and systematicity (\cref{sub_sec:system}), going test by test. For all experiments, the distribution between the two classes is perfectly balanced. Statistics on the diversity of the generated graphs with respect to relation paths can be found in \cref{tab:data_statistics}. Details on the selection of hyperparameters can be found in \cref{sec:appendix_hyperparams}.

\subsection{Substitutivity}\label{sub_sec:subst}
In our first test, we evaluate to what extent the five model configurations can answer questions targeting substitutivity, as defined in \cref{sec:overview_motivation}. Success at this task requires recognizing the mapping between the presence of certain relations in the graph and phrases in the natural language questions. Substitutivity is the most basic setting of our experiments and the results on this task demonstrate the general difficulty of the task.

\paragraph{Test design}
We generate 12\,000 question--graph pairs using our generation procedure. We do this for 2, 3 and 4-hops. Each experiment targets one of the hop lengths, i.e., both training and evaluating on the same number of hops. For all experiments, we use 10\,000 samples for training and 2\,000 for testing. All results are reported using mean accuracy and standard deviation from five randomly seeded runs.

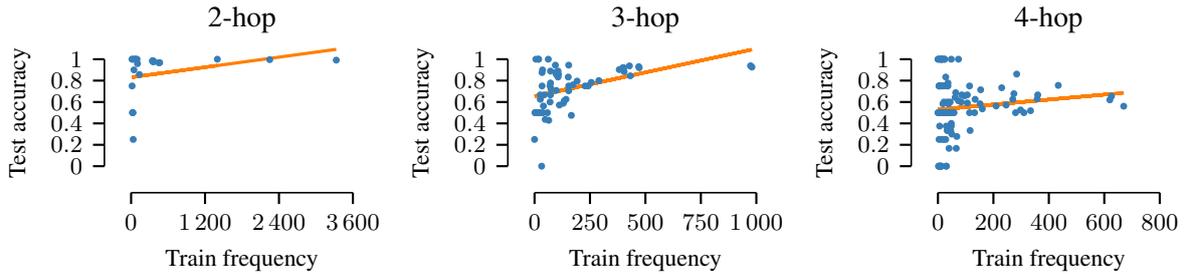
\begin{figure*}[htp]
\centering
\null%
\hfill%
\foreach \hop/\xmax/\xstep in {2/3600/1200,3/1000/250,4/800/200}{%
    \begin{tikzpicture}
    \begin{axis}[modern, width=45mm, height=30mm, ymin=0, ymax=1, ytick distance=0.2, xmin=0, xmax=\xmax, xtick distance=\xstep, clip=false, xlabel={Train frequency}, ylabel={Test accuracy}, ticklabel style={font=\footnotesize}, label style={font=\small}, title={\hop-hop}]
            \addplot[points] table {figures/\hop_regplot.dat};
            \addplot[regression] table[x=x, y={create col/linear regression={y=y}}] {figures/\hop_regplot.dat};
    \end{axis}
    \end{tikzpicture}%
    \hfill%
}%
\null
\caption{Relationship between frequency of a $k$-hop relation and the performance of the Disjoint model on these during testing.
Each blue point corresponds to a tuple of $k$ relations for $k=1,2,3$.
The orange lines are the linear regressions for each setting showing positive correlations between frequencies of $k$-hop relations and performances of the Disjoint model.}
\label{fig:subst_performance}
\end{figure*}

\paragraph{Results} The results of the substitutivity tests can be seen in \cref{tab:results_subst}. There is a strong generalization in the 2-hop case for all models. In the 3-hop case, performance drops significantly and the variance also increases for most models. There is a further loss in generalization when moving to the 4-hop case, where there is little difference between the models. \cref{fig:subst_performance} shows that there is a positive correlation between test accuracy on a specific question type and the frequency of the corresponding relation path in the training data for the 3-hop case. For the 2-hop and 4-hop cases, where the performance is closer to perfect and random classification, respectively, the relationship flattens. This aligns with the increase in unique combinations of relation paths as the number of hops increases, as shown in \cref{tab:data_statistics}. For 2-hops there is either a high number of examples per relation path combination, which enables learning, or very few, which makes the performance on these relation paths negligible on the overall accuracy. For 4-hops, on the other hand, there is too much of a long tail on the distribution of number of samples per combination, making learning difficult. For 3-hops, however, there is a balance, which is reflected in the results. See \cref{fig:dist_of_rel} in \cref{sec:appendix_dataset_generation} for an illustration of the relative percentage of each unique combination to the total for each hop.

These results show that the models struggle to generalize to the long tail of the training distribution, which is a known attribute of neural networks. However, single relation types are reused extensively throughout our training data, as there are only 10 unique relations of length 1. This means that the distribution only has a long tail if one considers each 4-hop combination to be a single unique type, as opposed to being composed of four individual relations that are individually high-frequent. We argue that these results indicate that these models do not create internal representations of each 1-hop path together with a function for composing them into compounded expressions, but rather learn individual representations for each compound. This also explains why the performance drops as the complexity increases and the number of samples per relation decreases, making such a strategy less viable. This explanation is further explored in the systematicity tests. We also note that there is little difference between the models, indicating that an increase in model complexity does not add any benefits to this task. 

\begin{table}[]
\centering
\footnotesize
    \begin{tabular}{@{}l@{\hspace{2em}}ccc@{}}
    \toprule
    \textbf{Model} & \textbf{2-hop} & \textbf{3-hop}  & \textbf{4-hop} \\
    \midrule
    Baseline & $48.88^{\pm1.00}$ & $49.8^{\pm0.64}$ & $49.36^{\pm0.44}$  \\
    Disjoint & $98.17^{\pm0.39}$ & $83.08^{\pm0.96}$ & $67.86^{\pm1.89}$ \\
    Grounded & $92.46^{\pm0.49}$ & $84.91^{\pm0.38}$ & $69.32^{\pm1.96}$ \\
    Unidirectional & $91.78^{\pm0.72}$ & $70.88^{\pm3.53}$ & $62.0^{\pm3.97}$ \\
    Bidirectional & $92.44^{\pm0.23}$ & $73.66^{\pm1.04}$ & $63.24^{\pm3.36}$ \\
    \bottomrule
    \end{tabular}
    \caption{Substitutivity results.}
    \label{tab:results_subst}
\end{table}

\subsection{Productivity}\label{sub_sec:prod}
In our next test, we evaluate to what extent models can answer $k$-hop questions when trained on $j$-hops, with $k \neq j$. We define experiments requiring \textit{extrapolation} and \textit{interpolation} for different values of $k$ and $j$. Success at this task requires generalization to sequences with a different length than those seen during training.

\paragraph{Test design}
In our first experiment, we use an even amount of 2 and 3-hop questions for training while using 4-hop questions for testing, targeting extrapolation. In our second experiment, we target interpolation by training on an even amount of 2 and 4-hop questions while testing on 3-hops. For our last experiment, we target extrapolation to lower values, using a training split with an even amount of 3 and 4-hop questions while testing on 2-hops. For all three experiments, we generate 10\,000 samples for training and 2\,000 for testing.

\paragraph{Results}
The results of the productivity tests can be seen in \cref{tab:results_productivity}. In general, we observe that all models struggle to extrapolate to both shorter and longer hops. We observe some generalization for the interpolation test, where the model has seen sequences both shorter and longer during training, but the overall absolute performance is still poor. This aligns with previous work, such as \citet{lake2018generalization} and \citet{hupkes2020compositionality}. We observe a tiny increase in performance with the more complex models compared to the simpler, disjoint model, but given the overall low scores and the variance, we do not think these results are significant. This indicates that adding a more complex interaction between the encoders does not provide any added benefit for extrapolation or interpolation.

\begin{table}[]
\centering
\footnotesize
    \begin{tabular}{@{}l@{\hspace{2em}}r@{}}
    \toprule
    \textbf{Experiment} & \textbf{Acc.}  \\
    \midrule
    \textsc{Train: 2 and 3-hop. Test: 4-hop}& \\
    \hspace{1em}Baseline & $49.92^{\pm0.82}$ \\
    \hspace{1em}Disjoint & $59.42^{\pm2.07}$  \\
    \hspace{1em}Grounded & $61.79^{\pm0.52}$   \\
    \hspace{1em}Unidirectional & $62.40^{\pm1.02}$   \\
    \hspace{1em}Bidirectional & $62.05^{\pm0.91}$   \\[0.5em]
    \textsc{Train: 2 and 4-hop. Test: 3-hop}& \\
    \hspace{1em}Baseline & $51.19^{\pm0.31}$ \\
    \hspace{1em}Disjoint & $63.87^{\pm2.24}$ \\
    \hspace{1em}Grounded & $66.10^{\pm2.02}$   \\
    \hspace{1em}Unidirectional & $66.65^{\pm2.24}$   \\
    \hspace{1em}Bidirectional & $66.29^{\pm1.25}$   \\[0.5em]
    \textsc{Train: 3 and 4-hop. Test: 2-hop}& \\
    \hspace{1em}Baseline & $50.98^{\pm0.25}$ \\
    \hspace{1em}Disjoint & $60.34^{\pm3.25}$  \\
    \hspace{1em}Grounded & $62.32^{\pm1.98}$   \\
    \hspace{1em}Unidirectional & $64.04^{\pm4.47}$   \\
    \hspace{1em}Bidirectional & $63.03^{\pm3.76}$   \\
    \bottomrule
    \end{tabular}
    \caption{Productivity results.}
    \label{tab:results_productivity}
\end{table}

\subsection{Systematicity}\label{sub_sec:system}
In our last test, we evaluate to what extent our models can generalize systematically. This means that all unique 1-hop relations are seen during training in some $k$-hop combination, but in the test data, we only include unseen combinations. For example, during training the models might see samples using the relations \textsc{takesCourse} and \textsc{hasTeachingAssistant}, and the 2-hop combination \textsc{[takesCourse, hasTeachingAssistant]}, but in the test we target \textsc{[hasTeachingAssistant, takesCourse]}. Success at this task requires the model to learn that a 3-hop question, for example, consists of three individual parts that are combined, as opposed to encoding the entire sequence as one entity. 

\paragraph{Test design}
With $k$ being 2, 3, or 4, we generate question--graph pairs with length $k$. Then, for each $k$, we randomly select a fraction of the available $k$-length combinations of relations as our test combinations. We then separate the generated pairs into two parts: those who use the test combinations and those who do not. The former is saved as the test set. We note that the use of test combinations is not limited to the relations on the path between the head and tail entities targeted by the question, but \emph{any} $k$-hop path that exists between these entities.
For the training set, we split the remaining possible combinations into five chunks and collect all samples that use these combinations. This leaves us with five different folds of the training set. Models are then trained on each fold independently, but evaluated on the same test set. We do this to control for the fact that some combinations of relations might have a similar distribution of samples as the relations picked out for the test set, making them better fits for that particular split. 

As this tests out-of-distribution generalization, we argue that folding is better suited for finding the true performance and variance of the models than the more straightforward evaluation method used for our substitutivity and productivity experiments. The results are reported as the mean accuracy over all five folds.

\paragraph{Results}
The results of the systematicity tests can be seen in \cref{tab:results_systematicity}. As with the productivity tests, we see that there is no consistent display of generalization for any of the models across hops. We see, however, that for the 3-hops setting the two simpler models achieve some success on the task. As these models had the highest performance on the substitutivity test, we take this to indicate that additional architectural complexity does not provide any performance benefit for this type of generalization. We also note that it is possible that these experiments are sensible to the hyperparameters, and that the two simpler models happened to have a better fit for the 3-hop setting. We did not do a specific hyperparameter search per model for the systematicity experiments but relied on those from the initial hyperparameter search. 

Overall, these results might come from under-sampling, or it might be the case, as is reported in \citet{hupkes2020compositionality}, that the models are not able to construct useful representations of smaller syntactic units. We argue that our results further support the intuition from the substitutivity experiments: the models do not learn to create representations that correspond to each 1-hop relation, together with a function for combining them, but rather rely on memorization of whole combinations of relation types. For short-tailed distributions and some settings of hyperparameters this works, but it is not a robust solution to the general problem of matching the text questions to certain patterns in the graphs, illustrated by the significant drop in performance in these results.

We also see adjacent work on assessing compositional generalizations in LLMs to provide possible explanations for our results. \citet{dziri2024faith} hypothesize that transformer-based models try to solve a highly compositional problem, like ours, by reducing it into linearized path matching. If this is the case, models do not learn the compositional rule needed to solve the task. If the samples from the test distribution only resemble the training distribution \textit{superficially}, this pattern matching approach is not a viable solution for out-of-distribution generalization in both our productivity and systematicity experiments. In these experiments, the mapping from the input samples to output labels cannot be based on surface-form frequency, due to our balanced sampling, which could then explain the collapse in performance we are observing for these experiments. 

\begin{table}[]
\centering
\footnotesize
    \begin{tabular}{@{}l@{\hspace{2em}}ccc@{}}
    \toprule
    \textbf{Model} & \textbf{2-hop} & \textbf{3-hop}  & \textbf{4-hop} \\
    \midrule
    Baseline & $49.50^{\pm0.24}$ & $49.40^{\pm0.33}$ & $49.92^{\pm1.01}$  \\
    Disjoint & $53.62^{\pm8.39}$ & $74.05^{\pm3.42}$ & $62.18^{\pm1.20}$ \\
    Grounded & $60.02^{\pm3.22}$ & $75.68^{\pm3.27}$ & $64.96^{\pm1.39}$ \\
    Unidir. & $64.24^{\pm0.75}$ & $57.12^{\pm1.81}$ & $64.10^{\pm3.62}$ \\
    Bidir. & $64.15^{\pm1.10}$ & $60.37^{\pm2.31}$ & $63.72^{\pm3.63}$ \\
    \bottomrule
    \end{tabular}
    \caption{Systematicity results.}
    \label{tab:results_systematicity}
\end{table}

\section{Previous work}
There have been numerous works on compositional generalization in neural networks \citep{andreas2018measuring, lake2018generalization, kim-linzen-2020-cogs, hupkes2020compositionality, dankers-etal-2022-transformer, lake2023comp, dziri2024faith}. Most similar to our work is \citet{gu2021beyond}, which develops a large-scale dataset targeting aspects of compositional generalization using question answering over Wikidata. Both this work and similar efforts reformulate the problem of integrating the KG to a seq2seq task where the questions are first translated into an intermediate logical form, such as SPARQL, which can be executed over the KG \citep{shu-etal-2022-tiara}. Under the same framework, \citet{ravishankar-etal-2022-two} explicitly tests generalization to unseen combinations of relations in a similar fashion to our systematicity test. They too find that the performance tends to drop significantly for this setting. 

The mapping from natural language to logical form can also use techniques from semantic parsing, a topic where compositionality is of special interest \citep{lindemann-etal-2023-compositional}. Our work, however, is the first to target compositionality in a setup where the models learn representations of both text and graphs \textit{directly}, without any logical form as an intermediate step. In contrast to previous work on grounded language models \citep{lin-etal-2019-kagnet, zhang-etal-2020-grounded, yasunaga-etal-2021-qa, greaselm, kaur-etal-2022-lm, sun-etal-2022-jointlk}, our methods can also more precisely quantify the effect of combining LMs with GNNs, as our sampling procedure ensures that the encoders can not achieve above random performance on their own, a property which we believe is unique to our approach.

\section{Conclusion}
Research on compositional generalization in neural networks has a long history. To the best of our knowledge, we provide the first experimental evaluation of compositional generalization in grounded LMs, using question answering over KGs as a task. Our results indicate that these models \emph{i)} require large sample sizes and quickly lose generalization power as the complexity of the compositional structure increases, \emph{ii)} struggle with extrapolation and interpolation to sequences of lengths not seen during training, and \emph{iii)} cannot systematically generalize from seen base components to novel compositions given the sample sizes used in our experiments. We show this across a set of model architectures inspired by the literature. Our results are in line with previous work on compositional generalization in other domains and for other types of neural architectures. Overall, we believe our results show that grounded LMs do not display signs of structured reasoning over KGs and that there is still much to be done on the improvement of models that try to exploit external information stored as symbolic objects. This is the case not only for knowledge graphs but also for other types of data representations, such as tabular data. We hope our generation procedure and released datasets can help motivate future efforts at overcoming these challenges.

\section*{Limitations}

\paragraph{Models} Our models are inspired by previous work. However, since our experimental setup differs from the original settings these models were implemented to work in, we had to make multiple simplifications. Consequently, we are not able to conclude anything about the abilities of the \textit{specific} models from the literature, but rather about the general interaction approach they propose.

\paragraph{Question variety} The questions in our dataset are formulated in English, and although they have more linguistic variety than similar studies from semantic parsing, such as \citet{lake2018generalization} and \citet{hupkes2020compositionality}, we still use a rather limited version of the language, which is by design devoid of any metaphors, idioms, analogies and other constructs that would make compositionality more difficult to measure. It is not given that our results transfer to real language data. 

\section*{Acknowledgments}
We acknowledge Sigma2, Norway, for awarding this project access to the LUMI supercomputer, owned by the EuroHPC Joint Undertaking, hosted by CSC (Finland) and the LUMI consortium.

\bibliography{anthology,custom}

\begin{thebibliography}{34}
\expandafter\ifx\csname natexlab\endcsname\relax\def\natexlab#1{#1}\fi

\bibitem[{Andreas(2019)}]{andreas2018measuring}
Jacob Andreas. 2019.
\newblock \href {https://openreview.net/forum?id=HJz05o0qK7} {Measuring compositionality in representation learning}.
\newblock In \emph{International Conference on Learning Representations}.

\bibitem[{Churchland(1990)}]{churchland1990cognitive}
Paul~M. Churchland. 1990.
\newblock Cognitive activity in artificial neural netowrks.
\newblock \emph{An invitation to cognitive science: Thinking}, 3:199--229.

\bibitem[{Dankers et~al.(2022{\natexlab{a}})Dankers, Bruni, and Hupkes}]{dankers-etal-2022-paradox}
Verna Dankers, Elia Bruni, and Dieuwke Hupkes. 2022{\natexlab{a}}.
\newblock \href {https://doi.org/10.18653/v1/2022.acl-long.286} {The paradox of the compositionality of natural language: A neural machine translation case study}.
\newblock In \emph{Proceedings of the 60th Annual Meeting of the Association for Computational Linguistics (Volume 1: Long Papers)}, pages 4154--4175, Dublin, Ireland. Association for Computational Linguistics.

\bibitem[{Dankers et~al.(2022{\natexlab{b}})Dankers, Lucas, and Titov}]{dankers-etal-2022-transformer}
Verna Dankers, Christopher Lucas, and Ivan Titov. 2022{\natexlab{b}}.
\newblock \href {https://doi.org/10.18653/v1/2022.acl-long.252} {Can transformer be too compositional? analysing idiom processing in neural machine translation}.
\newblock In \emph{Proceedings of the 60th Annual Meeting of the Association for Computational Linguistics (Volume 1: Long Papers)}, pages 3608--3626, Dublin, Ireland. Association for Computational Linguistics.

\bibitem[{Devlin et~al.(2019)Devlin, Chang, Lee, and Toutanova}]{devlin-etal-2019-bert}
Jacob Devlin, Ming-Wei Chang, Kenton Lee, and Kristina Toutanova. 2019.
\newblock \href {https://doi.org/10.18653/v1/N19-1423} {{BERT}: Pre-training of deep bidirectional transformers for language understanding}.
\newblock In \emph{Proceedings of the 2019 Conference of the North {A}merican Chapter of the Association for Computational Linguistics: Human Language Technologies, Volume 1 (Long and Short Papers)}, pages 4171--4186, Minneapolis, Minnesota. Association for Computational Linguistics.

\bibitem[{Dziri et~al.(2024)Dziri, Lu, Sclar, Li, Jiang, Lin, Welleck, West, Bhagavatula, Le~Bras et~al.}]{dziri2024faith}
Nouha Dziri, Ximing Lu, Melanie Sclar, Xiang~Lorraine Li, Liwei Jiang, Bill~Yuchen Lin, Sean Welleck, Peter West, Chandra Bhagavatula, Ronan Le~Bras, et~al. 2024.
\newblock Faith and fate: Limits of transformers on compositionality.
\newblock \emph{Advances in Neural Information Processing Systems}, 36.

\bibitem[{Fodor and Pylyshyn(1988)}]{fodor1988connectionism}
Jerry~A Fodor and Zenon~W Pylyshyn. 1988.
\newblock Connectionism and cognitive architecture: A critical analysis.
\newblock \emph{Cognition}, 28(1-2):3--71.

\bibitem[{Gu et~al.(2021)Gu, Kase, Vanni, Sadler, Liang, Yan, and Su}]{gu2021beyond}
Yu~Gu, Sue Kase, Michelle Vanni, Brian Sadler, Percy Liang, Xifeng Yan, and Yu~Su. 2021.
\newblock Beyond iid: three levels of generalization for question answering on knowledge bases.
\newblock In \emph{Proceedings of the Web Conference 2021}, pages 3477--3488.

\bibitem[{Guo et~al.(2005)Guo, Pan, and Heflin}]{guo2005lubm}
Yuanbo Guo, Zhengxiang Pan, and Jeff Heflin. 2005.
\newblock Lubm: A benchmark for owl knowledge base systems.
\newblock \emph{Journal of Web Semantics}, 3(2-3):158--182.

\bibitem[{Hosseini et~al.(2022)Hosseini, Vani, Bahdanau, Sordoni, and Courville}]{hosseini-etal-2022-compositional}
Arian Hosseini, Ankit Vani, Dzmitry Bahdanau, Alessandro Sordoni, and Aaron Courville. 2022.
\newblock \href {https://doi.org/10.18653/v1/2022.blackboxnlp-1.22} {On the compositional generalization gap of in-context learning}.
\newblock In \emph{Proceedings of the Fifth BlackboxNLP Workshop on Analyzing and Interpreting Neural Networks for NLP}, pages 272--280, Abu Dhabi, United Arab Emirates (Hybrid). Association for Computational Linguistics.

\bibitem[{Hupkes et~al.(2020)Hupkes, Dankers, Mul, and Bruni}]{hupkes2020compositionality}
Dieuwke Hupkes, Verna Dankers, Mathijs Mul, and Elia Bruni. 2020.
\newblock Compositionality decomposed: How do neural networks generalise?
\newblock \emph{Journal of Artificial Intelligence Research}, 67:757--795.

\bibitem[{Kaur et~al.(2022)Kaur, Bhatia, Aggarwal, Bansal, and Krishnamurthy}]{kaur-etal-2022-lm}
Jivat Kaur, Sumit Bhatia, Milan Aggarwal, Rachit Bansal, and Balaji Krishnamurthy. 2022.
\newblock \href {https://doi.org/10.18653/v1/2022.findings-naacl.57} {{LM}-{CORE}: Language models with contextually relevant external knowledge}.
\newblock In \emph{Findings of the Association for Computational Linguistics: NAACL 2022}, pages 750--769, Seattle, United States. Association for Computational Linguistics.

\bibitem[{Keysers et~al.(2020)Keysers, Sch{\"a}rli, Scales, Buisman, Furrer, Kashubin, Momchev, Sinopalnikov, Stafiniak, Tihon, Tsarkov, Wang, van Zee, and Bousquet}]{keysers2020measuring}
Daniel Keysers, Nathanael Sch{\"a}rli, Nathan Scales, Hylke Buisman, Daniel Furrer, Sergii Kashubin, Nikola Momchev, Danila Sinopalnikov, Lukasz Stafiniak, Tibor Tihon, Dmitry Tsarkov, Xiao Wang, Marc van Zee, and Olivier Bousquet. 2020.
\newblock \href {https://openreview.net/forum?id=SygcCnNKwr} {Measuring compositional generalization: A comprehensive method on realistic data}.
\newblock In \emph{International Conference on Learning Representations}.

\bibitem[{Kim and Linzen(2020)}]{kim-linzen-2020-cogs}
Najoung Kim and Tal Linzen. 2020.
\newblock \href {https://doi.org/10.18653/v1/2020.emnlp-main.731} {{COGS}: A compositional generalization challenge based on semantic interpretation}.
\newblock In \emph{Proceedings of the 2020 Conference on Empirical Methods in Natural Language Processing (EMNLP)}, pages 9087--9105, Online. Association for Computational Linguistics.

\bibitem[{Lake and Baroni(2018)}]{lake2018generalization}
Brenden~M. Lake and Marco Baroni. 2018.
\newblock \href {http://arxiv.org/abs/1711.00350} {Generalization without systematicity: On the compositional skills of sequence-to-sequence recurrent networks}.

\bibitem[{Lake and Baroni(2023)}]{lake2023comp}
Brenden~M. Lake and Marco Baroni. 2023.
\newblock \href {https://doi.org/10.1038/s41586-023-06668-3} {Human-like systematic generalization through a meta-learning neural network}.
\newblock \emph{Nature}, 623(7985):115--121.

\bibitem[{Lin et~al.(2019)Lin, Chen, Chen, and Ren}]{lin-etal-2019-kagnet}
Bill~Yuchen Lin, Xinyue Chen, Jamin Chen, and Xiang Ren. 2019.
\newblock \href {https://doi.org/10.18653/v1/D19-1282} {{K}ag{N}et: Knowledge-aware graph networks for commonsense reasoning}.
\newblock In \emph{Proceedings of the 2019 Conference on Empirical Methods in Natural Language Processing and the 9th International Joint Conference on Natural Language Processing (EMNLP-IJCNLP)}, pages 2829--2839, Hong Kong, China. Association for Computational Linguistics.

\bibitem[{Lindemann et~al.(2023)Lindemann, Koller, and Titov}]{lindemann-etal-2023-compositional}
Matthias Lindemann, Alexander Koller, and Ivan Titov. 2023.
\newblock \href {https://doi.org/10.18653/v1/2023.eacl-main.159} {Compositional generalisation with structured reordering and fertility layers}.
\newblock In \emph{Proceedings of the 17th Conference of the European Chapter of the Association for Computational Linguistics}, pages 2172--2186, Dubrovnik, Croatia. Association for Computational Linguistics.

\bibitem[{Pagin and Westerst{\aa}hl(2010)}]{pagin2010compositionality}
Peter Pagin and Dag Westerst{\aa}hl. 2010.
\newblock Compositionality i: Definitions and variants.
\newblock \emph{Philosophy Compass}, 5(3):250--264.

\bibitem[{Partee et~al.(1995)}]{partee1995lexical}
Barbara Partee et~al. 1995.
\newblock Lexical semantics and compositionality.
\newblock \emph{An invitation to cognitive science: Language}, 1:311--360.

\bibitem[{Ram et~al.(2023)Ram, Klinger, and Gray}]{ram2023how}
Parikshit Ram, Tim Klinger, and Alexander~G. Gray. 2023.
\newblock \href {https://openreview.net/forum?id=OImyRhNLv3} {How compositional is a model?}
\newblock In \emph{International Joint Conference on Artificial Intelligence 2023 Workshop on Knowledge-Based Compositional Generalization}.

\bibitem[{Ravishankar et~al.(2022)Ravishankar, Thai, Abdelaziz, Mihindukulasooriya, Naseem, Kapanipathi, Rossiello, and Fokoue}]{ravishankar-etal-2022-two}
Srinivas Ravishankar, Dung Thai, Ibrahim Abdelaziz, Nandana Mihindukulasooriya, Tahira Naseem, Pavan Kapanipathi, Gaetano Rossiello, and Achille Fokoue. 2022.
\newblock \href {https://doi.org/10.18653/v1/2022.findings-emnlp.408} {A two-stage approach towards generalization in knowledge base question answering}.
\newblock In \emph{Findings of the Association for Computational Linguistics: EMNLP 2022}, pages 5571--5580, Abu Dhabi, United Arab Emirates. Association for Computational Linguistics.

\bibitem[{Schlichtkrull et~al.(2018)Schlichtkrull, Kipf, Bloem, Van Den~Berg, Titov, and Welling}]{schlichtkrull2018modeling}
Michael Schlichtkrull, Thomas~N Kipf, Peter Bloem, Rianne Van Den~Berg, Ivan Titov, and Max Welling. 2018.
\newblock Modeling relational data with graph convolutional networks.
\newblock In \emph{The Semantic Web: 15th International Conference, ESWC 2018, Heraklion, Crete, Greece, June 3--7, 2018, Proceedings 15}, pages 593--607. Springer.

\bibitem[{Shu et~al.(2022)Shu, Yu, Li, Karlsson, Ma, Qu, and Lin}]{shu-etal-2022-tiara}
Yiheng Shu, Zhiwei Yu, Yuhan Li, B{\"o}rje Karlsson, Tingting Ma, Yuzhong Qu, and Chin-Yew Lin. 2022.
\newblock \href {https://doi.org/10.18653/v1/2022.emnlp-main.555} {{TIARA}: Multi-grained retrieval for robust question answering over large knowledge base}.
\newblock In \emph{Proceedings of the 2022 Conference on Empirical Methods in Natural Language Processing}, pages 8108--8121, Abu Dhabi, United Arab Emirates. Association for Computational Linguistics.

\bibitem[{Smolensky(1987)}]{smolensky1987connectionist}
Paul Smolensky. 1987.
\newblock Connectionist ai, symbolic ai, and the brain.
\newblock \emph{Artificial Intelligence Review}, 1(2):95--109.

\bibitem[{Sun et~al.(2022)Sun, Shi, Qi, and Zhang}]{sun-etal-2022-jointlk}
Yueqing Sun, Qi~Shi, Le~Qi, and Yu~Zhang. 2022.
\newblock \href {https://doi.org/10.18653/v1/2022.naacl-main.372} {{J}oint{LK}: Joint reasoning with language models and knowledge graphs for commonsense question answering}.
\newblock In \emph{Proceedings of the 2022 Conference of the North American Chapter of the Association for Computational Linguistics: Human Language Technologies}, pages 5049--5060, Seattle, United States. Association for Computational Linguistics.

\bibitem[{Szabó(2022)}]{sep-compositionality}
Zoltán~Gendler Szabó. 2022.
\newblock {Compositionality}.
\newblock In Edward~N. Zalta and Uri Nodelman, editors, \emph{The {Stanford} Encyclopedia of Philosophy}, {F}all 2022 edition. Metaphysics Research Lab, Stanford University.

\bibitem[{Wang et~al.(2019)Wang, Kapanipathi, Musa, Yu, Talamadupula, Abdelaziz, Chang, Fokoue, Makni, Mattei et~al.}]{wang2019improving}
Xiaoyan Wang, Pavan Kapanipathi, Ryan Musa, Mo~Yu, Kartik Talamadupula, Ibrahim Abdelaziz, Maria Chang, Achille Fokoue, Bassem Makni, Nicholas Mattei, et~al. 2019.
\newblock Improving natural language inference using external knowledge in the science questions domain.
\newblock In \emph{Proceedings of the AAAI Conference on Artificial Intelligence}, volume~33, pages 7208--7215.

\bibitem[{Wold et~al.(2023)Wold, {\O}vrelid, and Velldal}]{wold-etal-2023-text}
Sondre Wold, Lilja {\O}vrelid, and Erik Velldal. 2023.
\newblock \href {https://doi.org/10.18653/v1/2023.matching-1.1} {Text-to-{KG} alignment: Comparing current methods on classification tasks}.
\newblock In \emph{Proceedings of the First Workshop on Matching From Unstructured and Structured Data (MATCHING 2023)}, pages 1--13, Toronto, ON, Canada. Association for Computational Linguistics.

\bibitem[{Yasunaga et~al.(2022)Yasunaga, Bosselut, Ren, Zhang, Manning, Liang, and Leskovec}]{yasunaga2022deep}
Michihiro Yasunaga, Antoine Bosselut, Hongyu Ren, Xikun Zhang, Christopher~D Manning, Percy Liang, and Jure Leskovec. 2022.
\newblock \href {https://openreview.net/forum?id=4NpoSrT8uU-} {Deep bidirectional language-knowledge graph pretraining}.
\newblock In \emph{Advances in Neural Information Processing Systems}.

\bibitem[{Yasunaga et~al.(2021)Yasunaga, Ren, Bosselut, Liang, and Leskovec}]{yasunaga-etal-2021-qa}
Michihiro Yasunaga, Hongyu Ren, Antoine Bosselut, Percy Liang, and Jure Leskovec. 2021.
\newblock \href {https://doi.org/10.18653/v1/2021.naacl-main.45} {{QA}-{GNN}: Reasoning with language models and knowledge graphs for question answering}.
\newblock In \emph{Proceedings of the 2021 Conference of the North American Chapter of the Association for Computational Linguistics: Human Language Technologies}, pages 535--546, Online. Association for Computational Linguistics.

\bibitem[{Yu et~al.(2022)Yu, Zhu, Yang, and Zeng}]{Yu_Zhu_Yang_Zeng_2022}
Donghan Yu, Chenguang Zhu, Yiming Yang, and Michael Zeng. 2022.
\newblock \href {https://doi.org/10.1609/aaai.v36i10.21417} {Jaket: Joint pre-training of knowledge graph and language understanding}.
\newblock \emph{Proceedings of the AAAI Conference on Artificial Intelligence}, 36(10):11630--11638.

\bibitem[{Zhang et~al.(2020)Zhang, Liu, Xiong, and Liu}]{zhang-etal-2020-grounded}
Houyu Zhang, Zhenghao Liu, Chenyan Xiong, and Zhiyuan Liu. 2020.
\newblock \href {https://doi.org/10.18653/v1/2020.acl-main.184} {Grounded conversation generation as guided traverses in commonsense knowledge graphs}.
\newblock In \emph{Proceedings of the 58th Annual Meeting of the Association for Computational Linguistics}, pages 2031--2043, Online. Association for Computational Linguistics.

\bibitem[{Zhang et~al.(2022)Zhang, Bosselut, Yasunaga, Ren, Liang, Manning, and Leskovec}]{greaselm}
Xikun Zhang, Antoine Bosselut, Michihiro Yasunaga, Hongyu Ren, Percy Liang, Christopher~D Manning, and Jure Leskovec. 2022.
\newblock \href {https://openreview.net/forum?id=41e9o6cQPj} {Grease{LM}: Graph {REAS}oning enhanced language models}.
\newblock In \emph{International Conference on Learning Representations}.

\end{thebibliography}

\clearpage
\onecolumn
\appendix

\setlength{\parindent}{0pt}
\setlength{\parskip}{0.7em}

\renewcommand{\arraystretch}{1.2}

\section{Dataset generation}
\label{sec:appendix_dataset_generation}

\subsection{LUBM}
We use the LUBM UBA to generate knowledge graphs on the fly.\footnote{\url{https://swat.cse.lehigh.edu/projects/lubm/}} We limit our graphs to the following entity types: \textit{Professor}, \textit{AssociateProfessor}, \textit{UndergraduateStudent}, \textit{GraduateStudent}, \textit{TeachingAssistant}, \textit{Publication}, \textit{Course}; and the following relations: \textsc{publicationAuthor}, \textsc{teacherOf}, \textsc{advisor}, \textsc{takesCourse}, \textsc{teachingAssistantOf}. We also add reverse edges with reverse relations for all of these to create more diverse question types in our final dataset, making $|\mathcal{R}_{LUBM}'| = 10$. We remove all literals. Using a sample size of approximately $14\,000$, we create graphs that have $\mu=685, \sigma=77.16$ nodes and $\mu=4949, \sigma=532$ edges after processing.

\subsection{Statistics}\label{sub_sec:data_statistics}
\begin{figure}[h]
\centering
\begin{tikzpicture}
\begin{axis}[modern, width=0.8\textwidth, height=8cm, xmin=0, xmax=200, xtick distance=20, ymin=0, ymax=0.35, ytick distance=0.05, clip=false, xlabel={Relation path}, ylabel={Train frequency}, tick label style={/pgf/number format/fixed}, legend style={draw=none}]
    \addplot+[no markers, thick, draw=pblue] table[x expr=\coordindex, y=f] {distribution_2hop.dat};
    \addplot+[no markers, thick, draw=porange] table[x expr=\coordindex, y=f] {distribution_3hop.dat};
    \addplot+[no markers, thick, draw=pgreen] table[x expr=\coordindex, y=f] {distribution_4hop.dat};
    \legend{2-hop, 3-hop, 4-hop}
\end{axis}
\end{tikzpicture}
\caption{The relative percentage of each unique relation path to the total for 2, 3, and 4-hops.}
\label{fig:dist_of_rel}
\end{figure}
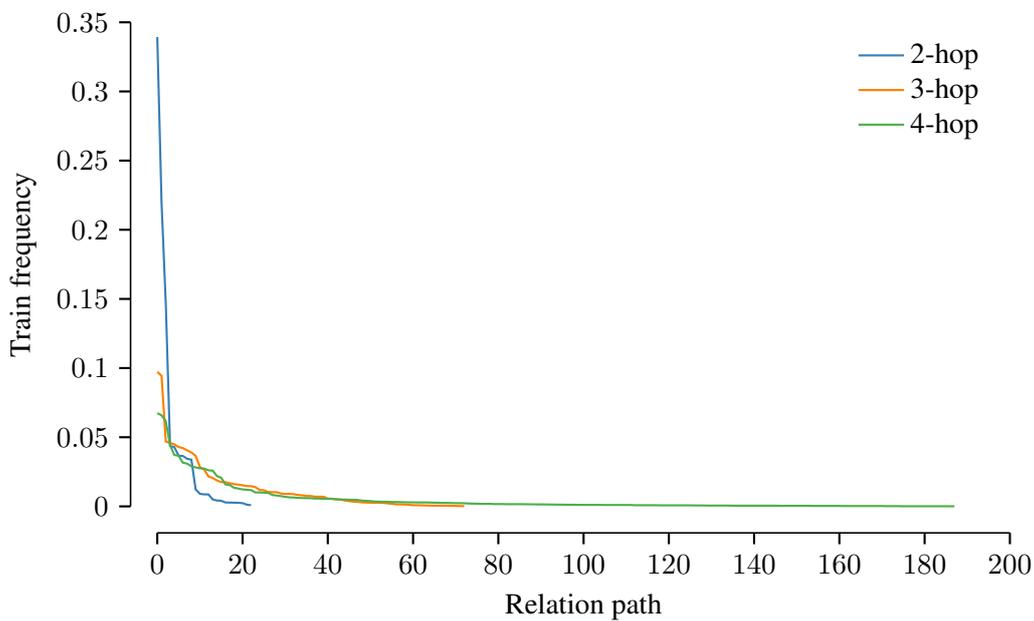
The number of possible combinations of relations on the edges in a $k$-hop path increases with the number of hops. As our sampling procedure uses $10$ distinct relations we have $10^k$ theoretical combinations of relations for any $k$-hop path. As our graphs are committed to the LUBM ontology the number of relations that hold between two entity types is much lower. 

\newpage
\subsection{Sampling procedure}
Given a directed multigraph $G = (\mathcal{V}, \mathcal{E})$ with labeled entities $v_i \in \mathcal{V}$ and labeled edges $(v_i, r, v_j) \in \mathcal{E}$ where $r \in \mathcal{R}$ is a relation, we sample a positive and negative example for a $k$-hop path using the following procedure:

\begin{enumerate}
    \item Sample two pairs of nodes ($s_+$ and $e_+$) and ($s_-$ and $e_-$) from $\mathcal{V}$ with uniform probability, making sure that $s_+ \neq s_-$ or $e_+ \neq e_-$, that $s_+$ and $s_-$ are of the same entity type, and that $e_+$ and $e_-$ are of the same entity type.
    \item Let $R_+$ be the set of all relation paths of length $k$ from $s_+$ to $e_+$
    \item Let $R_-$ be the set of all relation paths of length $k$ from $s_-$ to $e_-$
    \item Let $P_+$ be the set $R_+ \setminus R_-$
    \item Let $P_-$ be the set $R_- \setminus R_+$
    \item If $P_+ \neq \emptyset$ and $P_- \neq \emptyset$:
    \begin{enumerate}
        \item Sample a relation path $q$: $q \sim \mathcal{U}(P_+)$
        \item Return the triple $(q, s_+, e_+)$ as a positive example and $(q, s_-, e_-)$ as a negative example.
    \end{enumerate}
\end{enumerate}
This procedure returns a $k$-hop relation path, $q$, that holds between $s_+$ and $e_+$ but not between $s_-$ and $e_-$. However, we also make sure that there are paths that hold between $s_-$ and $e_-$ but not $s_+$ and $e_+$, ensuring symmetry.

In addition, we shuffle entities labels while preserving entity types before running this procedure since the LUBM generator introduces problematic patterns with its entity numbering.
For example, in the original graph, Professor1 is more likely to author Paper1 than Paper5.
By randomly exchanging the labels of Paper1 and Paper5, we ensure the models cannot exploit this regularity.

\section{Models}
\label{sec:appendix_hyperparams}

\subsection{Parameters}

All our combined models have a total of $\approx 141\,000\,000$ parameters, with the majority of them coming from the language model. Our baseline has $\approx 30\,000\,000$. 

\subsection{Computation infrastructure}
All the experiments ran on the LUMI supercomputer,\footnote{\url{https://www.lumi-supercomputer.eu/sustainable-future/}} using a single AMD MI250X GPU. Running all the experiments reported in the paper has a total computing budget of approximately 150 hours.

\subsection{Hyperparameters}

We tune the learning rate for the text encoder from $\{1 \times 10^{-6}, 1 \times 10^{-5}, 2 \times 10^{-5}, 3 \times 10^{-5}, 5 \times 10^{-5}\}$, the graph encoder module from $\{2 \times 10^{-4}, 5 \times 10^{-4}, 1 \times 10^{-3}, 2 \times 10^{-3}, 1 \times 10^{-5}\}$, and the MLP module from $\{1 \times 10^{-4}, 1 \times 10^{-5}\}$; the dropout value from $\{0.1, 0.2, 0.3\}$ and the epochs from $\{5, 8, 10\}$, with no early stopping. The hyperparameters are tuned on a development set targeting the substitutivity task with an equal mix of 2, 3, and 4-hop questions. We run 60 trials for each of the five model configurations and sample a random combination of these parameters for each trial, selecting the best combination for our experiments. We find that all models perform the best with no decay on the learning rate. 

\begin{table*}[h!]
\centering
\small
\begin{tabular}{@{}lccccc@{}}
\toprule
\textbf{Hyperparameter} & \textsc{Baseline} & \textsc{Disjoint} & \textsc{Grounded} & \textsc{Unidirectional} & \textsc{Bidirectional}  \\ \midrule
LM lr                 & - & 0.00001 & 0.00001 & 0.00001 & 0.00001         \\
GNN lr                & 0.00001 & 0.0005   & 0.0005 & 0.00005 & 0.0005          \\
MLP lr               & 0.00001 & 0.0001  & 0.0001 & 0.0001 & 0.0001          \\
Dropout               & 0.3     & 0.3     & 0.3 & 0.3 & 0.3          \\
Epochs                & 5      & 8     & 8 & 8 & 8 \\
\end{tabular} %
\caption{Variable training hyperparameters for all five model configurations}
\label{tab:hyperparams}
\end{table*}

\begin{table*}[h!]
\centering
\small
\begin{tabular}{@{}lc@{}}
\toprule
\textbf{Hyperparameter} & \text{Value}   \\ \midrule
Pre-trained LM          & bert-base-uncased \citep{devlin-etal-2019-bert}     \\
LM hidden size          & 768     \\
GNN hidden size         & 768       \\
GNN layers              & 4       \\
Batch size              & 16       \\
Sequence length         & 64        \\
Learning rate decay     & constant        \\
Optimizer               & AdamW         \\
\end{tabular} %
\caption{Constant training hyperparameters for all five model configurations}
\label{tab:static_hyperparams}
\end{table*}

\end{document}